# SignalGP-Lite: Event Driven Genetic Programming Library for Large-Scale Artificial Life Applications

**Matthew Andres Moreno**[1], **Santiago Rodriguez Papa**[1], **Alexander Lalejini**[2], **and Charles Ofria**[1]

**1** Michigan State University **2** University of Michigan

## Abstract

Event-driven genetic programming representations have been shown to outperform traditional imperative representations on interaction-intensive problems. The event-driven approach organizes genome content into modules that are triggered in response to environmental signals, simplifying simulation design and implementation. Existing work developing event-driven genetic programming methodology has largely used the SignalGP library, which caters to traditional program synthesis applications. The SignalGP-Lite library enables larger-scale artificial life experiments with streamlined agents by reducing control flow overhead and trading run-time flexibility for better performance due to compile-time configuration. Here, we report benchmarking experiments that show an 8x to 30x speedup. We also report solution quality equivalent to SignalGP on two benchmark problems originally developed to test the ability of evolved programs to respond to a large number of signals and to modulate signal response based on context.

## Summary

SignalGP is an existing event-driven genetic programming C++ library well-suited for interaction-heavy program synthesis problems. SignalGP was instrumental in introducing the event-driven genetic programming paradigm where program modules are triggered in response to signals from the environment.

Unlike the traditional imperative genetic programming paradigm, where a single chain of execution directly manages every aspect of the program, event-driven genetic programs trigger event handlers (i.e., program modules) in response to signals that are generated internally, externally from other agents, or externally from the environment. Event-driven representation outperforms traditional imperative genetic programming on interaction intensive problems where the programs must handle inputs from the environment or other organisms, as is the case in some traditional genetic programming contexts and many artificial life simulations.

SignalGP-Lite is a C++ library for event-driven genetic programming. In comparison to SignalGP, which is intended for general event-driven genetic programming, SignalGP-Lite is tailored for use in artificial life experiments.

Here, we present the compute time and solution quality comparison of SignalGP-Lite and SignalGP on some scenarios where the original implementation has the edge.

In "Execution Speed Benchmarking," we report compute times for both SignalGP and SignalGP-Lite using synthetic benchmarks—benchmarks that designed with reproducibility and accuracy in mind, but that might not reflect real-world problems. In "Test Problem Benchmarking," we compare solution quality of SignalGP and SignalGP-Lite on synthetic genetic programming problems designed to test responsivity and plasticity.

## Statement of need

Despite being able to simulate evolution with much faster generational turnover than is possible in biological experiments (Ofria & Wilke, 2004), the scale of artificial life populations is profoundly limited by available computational resources (Matthew A. Moreno, 2020). Large population sizes are essential to studying fundamental evolutionary phenomena such as ecologies, the transition to multicelularity, and rare events. In conjunction with parallel and distributed computing, computational efficiency is crucial to achieving larger-scale artificial life situations.

SignalGP-Lite is the first implementation of SignalGP accessible to the general public. Its simplified API is complemented by extensive documentation, continuous integration tests, and benchmarks that measure both real-world workloads as well as individual instructions. In comparison to SignalGP—which was designed to target generic genetic programming problems—SignalGP-Lite fills a niche for interaction-heavy genetic programming applications that can tolerate less runtime configuration flexibility, and pared-back control flow. SignalGP-Lite is designed with artificial life experiments in mind, where simulation parameters need not change during execution and a more rudimentary approach to control flow can often be tolerated.

The library has enabled order-of-magnitude scale-up of existing artificial life experiments study-ing the evolution of multicelularity; we anticipate it will also enable novel work in other artificial life and genetic programming contexts.

## Execution Speed Benchmarking

We performed a set of microbenchmarks—a type of synthetic benchmark that measures individual functions—to quantify the effectiveness of SignalGP-Lite's optimizations in accelerating evaluation of event-driven genetic programs.

Hardware caching size profoundly affects memory access time, which is key to computational performance (Skadron et al., 1999). In order to determine the relative performance of SignalGP and SignalGP-Lite at different cache levels, we benchmarked over different orders of magnitude of memory load by varying the number of virtual CPUs (agent counts) between from 1 and 32768 (Supplementary Table 1).

We performed five microbenchmark experiments, reported below, to isolate how specific aspects of the library design influenced performance. Analysis below focuses on wall time speedup. However, supplementary Figure 6 shows raw wall-clock timings for these experiments.

### control

The control involves importing the library to benchmark, initializing agents, and then measuring the execution time of an empty loop. This experiment verifies the validity of our benchmarking process. The 1x wall speedup (Figure 3) confirms that further results are not inadvertently skewed by our experimental apparatus.

### nop

A program consisting of 100 `nop` instructions is randomly generated. None of these instructions advance the pRNG engine. This benchmarks the instruction directly, as it is the only call measured inside the benchmarking loop. With this approach, the relative performance impact of SignalGP-Lite's byte-code interpreter can be compared to SignalGP's lambda-based instructions.

We observe an 8x to 30x speedup under SignalGP-Lite (Figure 3). The greatest speedup occurred at a relatively light memory footprint of 1024 agents.

### arithmetic

A program consisting of 100 randomly-chosen arithmetic instructions (`add`, `subtract`, `multiply`, and `divide`) is generated. This measures the performance impact of SignalGP-Lite's fixed-length array registers compared to SignalGP's variable-length vector registers. This compile-time optimization streamlines register access at the cost of the ability to change the number of registers on the fly. Since our aim is to only measure the performance effect of this optimization, no `nop` instructions are present in the generated program.

Figure 3 shows that incorporating this trade-off increases speedup to 20x to 50x. The greatest speedup increase occurred at a relatively light memory footprint of 1024 agents.

### complete

The complete benchmark adds control flow instructions to the prior benchmarks' instruction set. Bitwise and logical operators, comparison instructions, and RNG operations, are also included. From this complete instruction set, a 100-instruction program is randomly generated.

The main goal of this benchmark is to determine the performance impact of omitting a function stack and implementing inner loops and conditionals in terms of `jump` instructions instead of nested code blocks.

SignalGP-Lite's stripped-down control flow model increases speedup to 30x to 55x compared to vanilla SignalGP (Figure 3). The greatest speedup occurred at a light memory footprint of 32 agents.

### sans_regulation

Regulation operations allow SignalGP and SignalGP-Lite programs to adjust which program modules are expressed in response to environmental signals. Since this invalidates tag-match caches (which help lower the performance impact of tag-matching), we wanted to measure timings without regulation enabled.

This benchmark measures the complete instruction set with regulation-related instructions excluded.

As shown on Figure 3, this yields a 35x to 47x speed-up with respect to SignalGP. The greatest speedup occurred at a light memory footprint of 32 agents. From this, we can conclude that SignalGP-Lite offers performance improvements even on simulations that do not heavily depend on regulation.

## Test Problem Benchmarking

In order to viably serve as a specialized alternative to the original SignalGP for certain artificial life applications, SignalGP-Lite must match SignalGP's performance on benchmarks measuring responsivity and plasticity (the ability of organisms to adapt to changes in their environment). To verify SignalGP-Lite's aptitude on these tests, we replicated two canonical SignalGP experiments, reported below (Lalejini et al., 2021; Lalejini & Ofria, 2018).

### Changing Environment Problem

The Changing Environment Problem dispatched K = 2, 4, 8, or 16 mutually-exclusive environmental signals with randomly generated labels. Organisms were tasked to respond to each signal with a unique response instruction (Lalejini & Ofria, 2018).

A total of 100 replicate populations of 100 individuals were evolved for up to 10,000 generations. Elite selection was used to choose the best-fit individual; roulette selection was used for the other 99. Figure 1 shows the number of generations elapsed before a full solution was found. SignalGP-Lite evolved full solutions to each problem within 3,500 updates in all 100 tested replicates.

In the K=16 case, we achieved a superior 100% signal reproduction rate compared to an average of 32% on the original SignalGP implementation (Lalejini & Ofria (2018) Figure 2). We suspect this improvement occurred due to differences in how mutation, tag matching, and program initialization were performed, rather than an intrinsic difference between the libraries.

### Contextual Signal Problem

The Contextual Signal Problem assesses the ability of evolving programs to maintain memory of previously encountered signals. In previous work, this problem was used to demonstrate an important use case of regulation instructions. To solve this problem, programs must remember an initial signal (i.e., its "context") in order to respond appropriately to a second signal (Lalejini et al., 2021).

We assigned each possible unordered input signal pair a unique response to then be performed by the organism. We tested with 16 input signal pairs and 4 output responses. Table 2 in Lalejini et al. (2021) enumerates these sequences and responses.

A total of 20 replicates were evolved for up to 10,000 generations using a 16-way lexicase selection scheme (Spector, 2012), with each of the input signal pairs serving as a test case. To evaluate each test case, programs were sent the first signal of each test case and given 128 virtual CPU cycles to process it. After this, their internal running modules were killed and the second signal was sent. After another 128 virtual CPU cycles, their response was recorded. In order to save resources and computing time, as soon as a replicate evolved a fully-correct solution, their evolution was halted. We excluded RNG operations from the instruction set to ensure that solutions were not reached by chance. Figure 2 shows the number of generations elapsed before a full solution was found.

SignalGP-Lite evolved full solutions in half as many generations compared to SignalGP when regulation was enabled. Moreover, fewer replicates failed to reach a full solution in 10000 generations under SignalGP-Lite. With regulation disabled, however, the performance of both libraries was similar. These results mean that SignalGP-Lite is a valid alternative to SignalGP when it comes to artificial life applications.

## Projects Using the Software

SignalGP-Lite is used in DISHTINY, a digital framework for studying organism multicelularity (Matthew Andres Moreno & Ofria, 2019).

## Figures

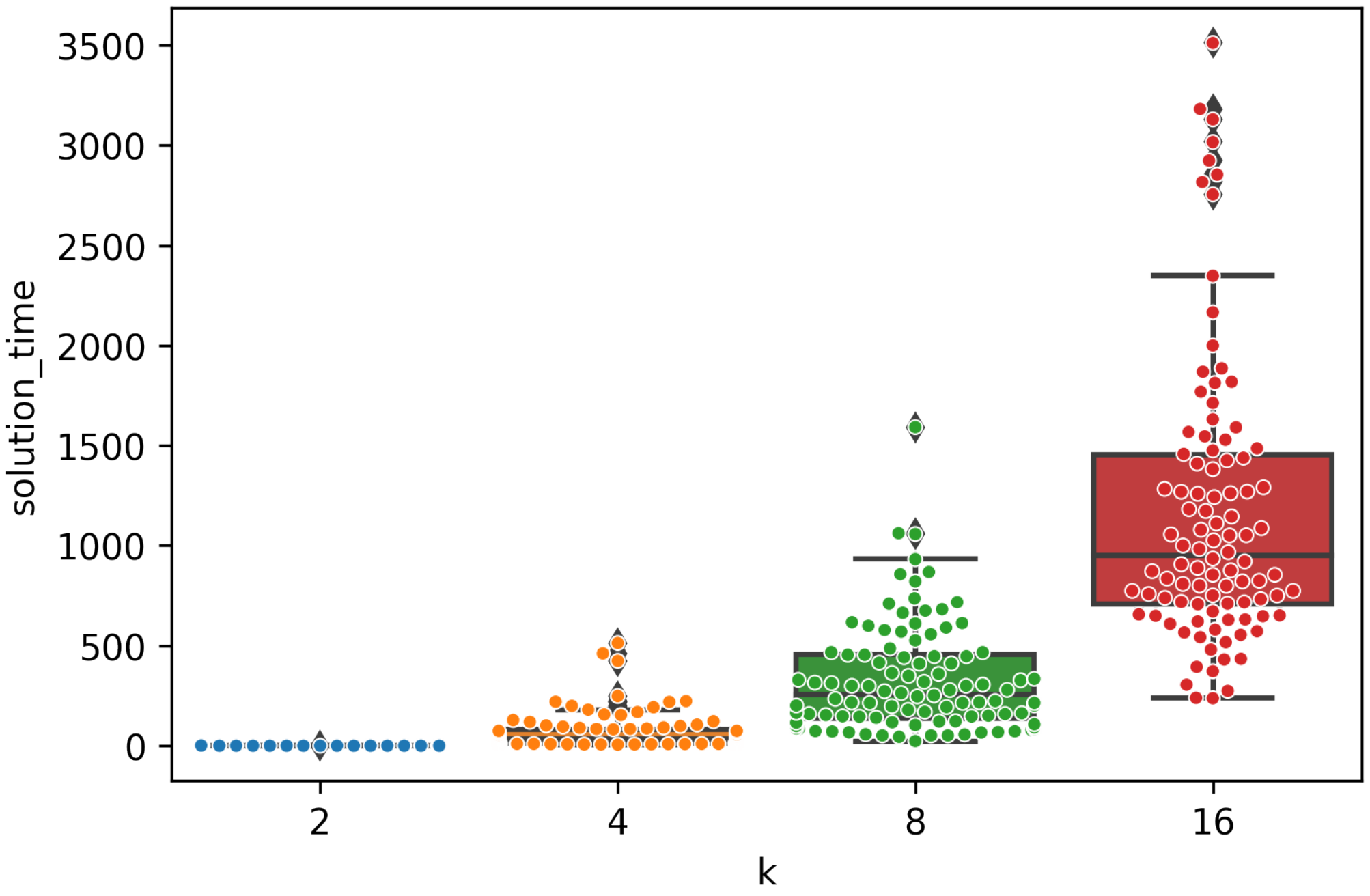

**Figure 1:** Number of generations elapsed before a perfect solution was observed on the Changing Enviroment problem. All replicates found a perfect solution.

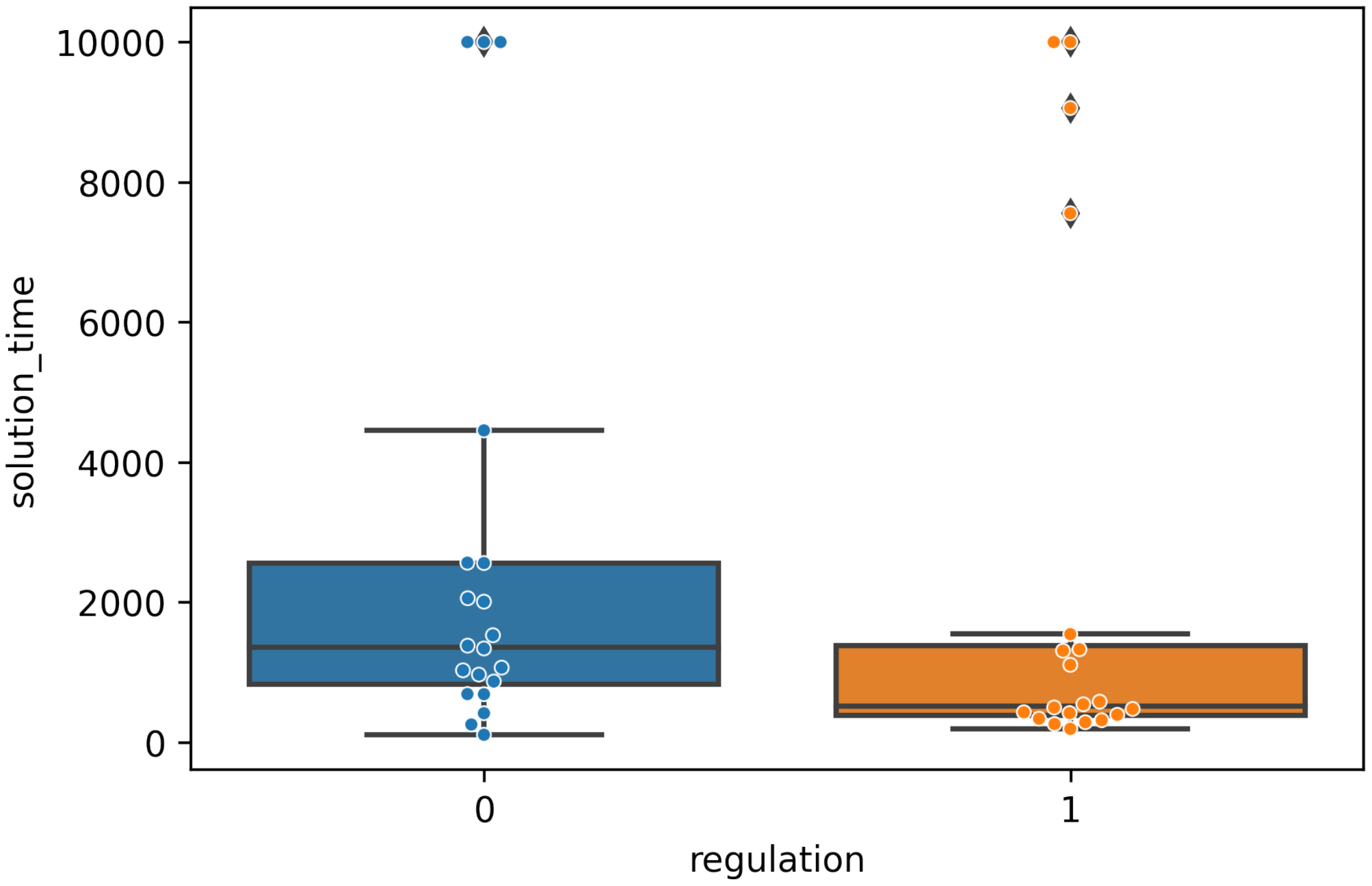

**Figure 2:** Number of generations elapsed before a perfect solution was observed on the Contextual Signal problem. Replicates that did not find a solution are on a dashed line at 10,000 generations.

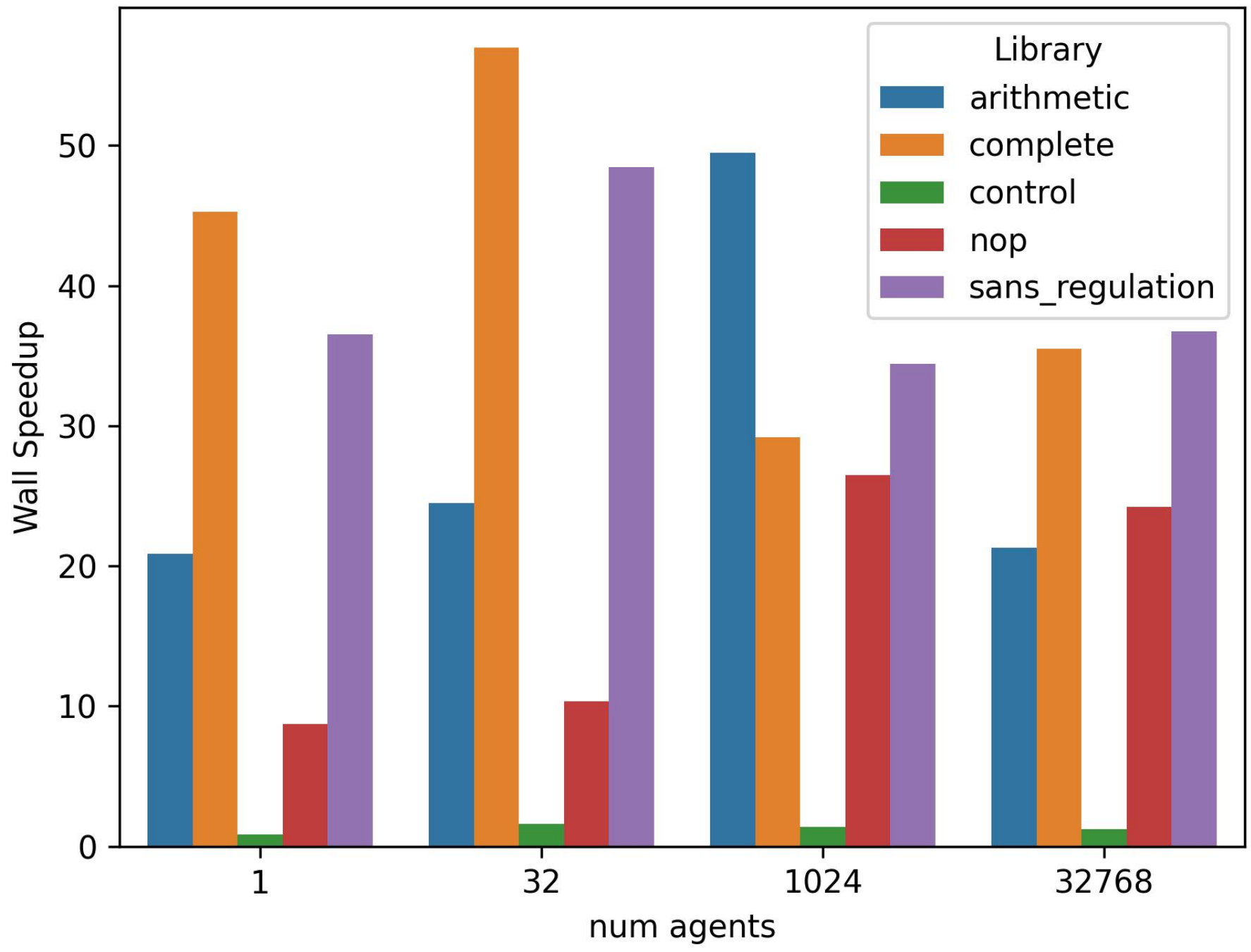

**Figure 3:** Benchmarking results of 20 replicates shown as a times-speedup of wall time. "Library" refers to the set of instructions tested (see Benchmarking Results section).

## Acknowledgements

This research was supported in part by NSF grants DEB-1655715 and DBI-0939454 as well as by Michigan State University through the computational resources provided by the Institute for Cyber-Enabled Research. This material is based upon work supported by the National Science Foundation Graduate Research Fellowship under Grant No. DGE-1424871, and by the Michigan State University BEACON Center Luminaries program. Any opinions, findings, and conclusions or recommendations expressed in this material are those of the author(s) and do not necessarily reflect the views of the National Science Foundation.

## Supplementary Material

All benchmarks reported in this section were performed using Google Benchmark version 1.5.2-1.

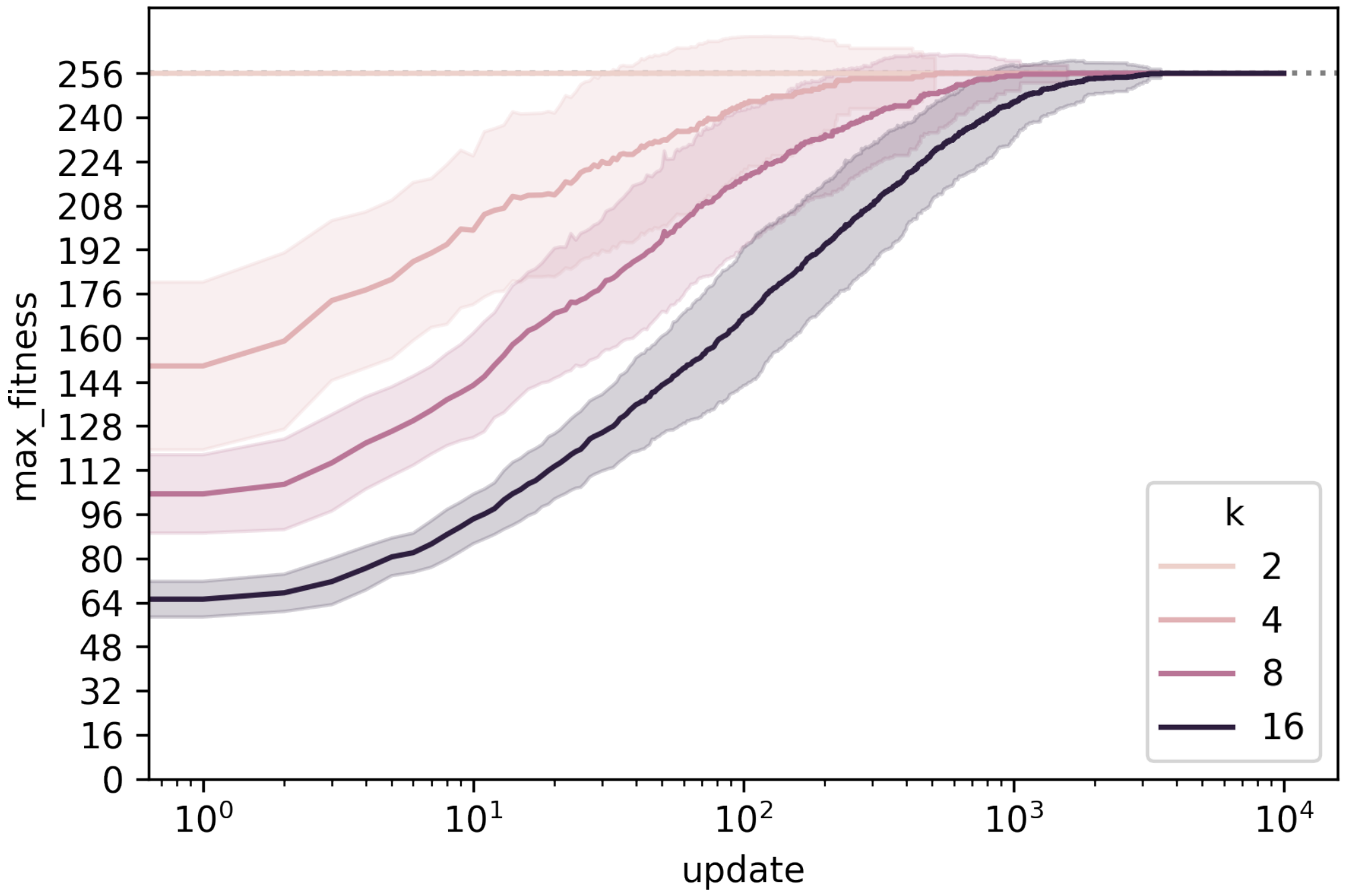

**Figure 4:** Maximum fitness wrt. updates, with standard deviation confidence intervals. This is because, due to large number of datapoints, computing 95% CI takes a non-insignificant amount of time.

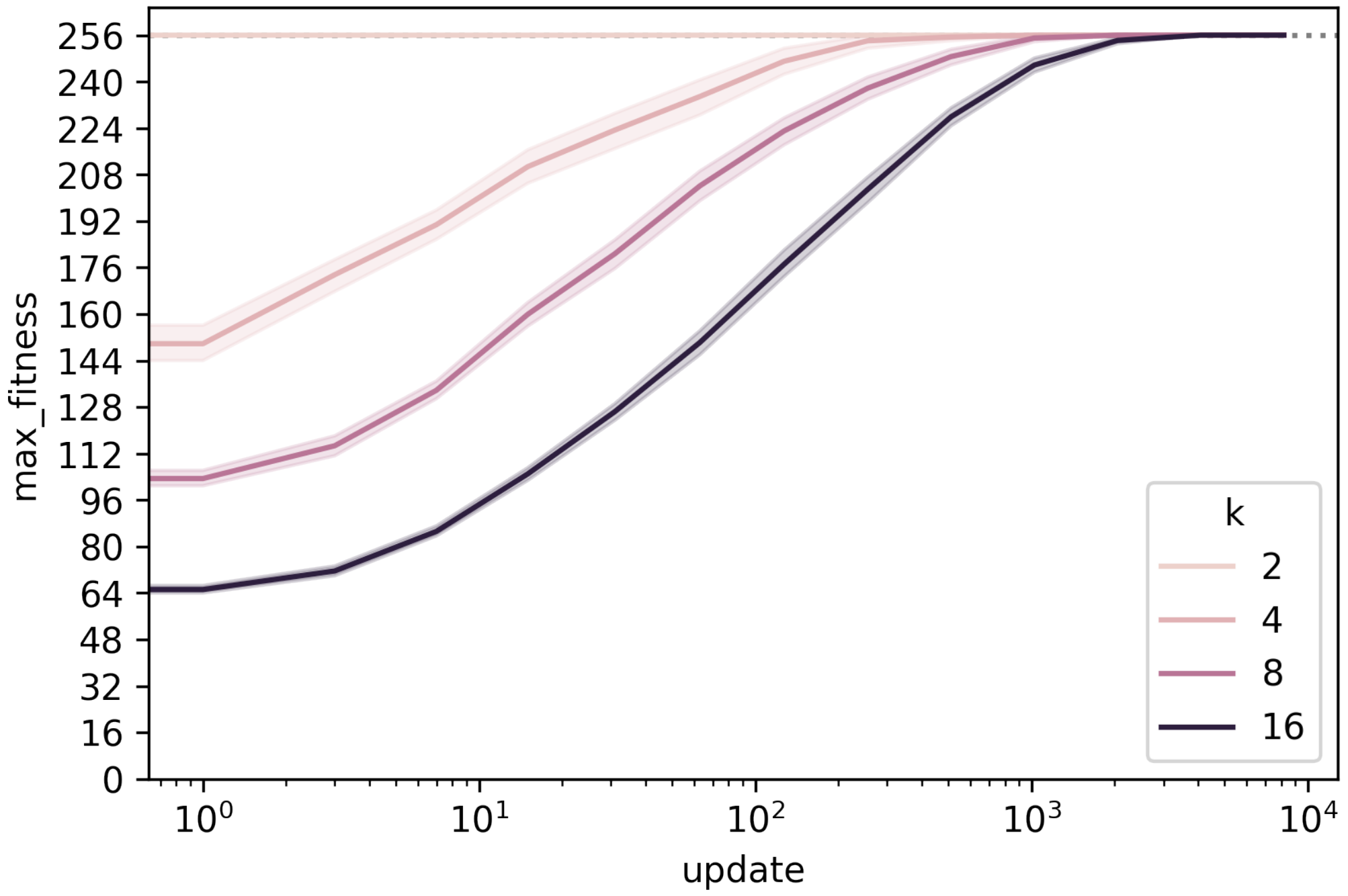

**Figure 5:** Filtered maximum fitness wrt. updates, with 95% confidence intervals. Data has been filtered logarithmically.

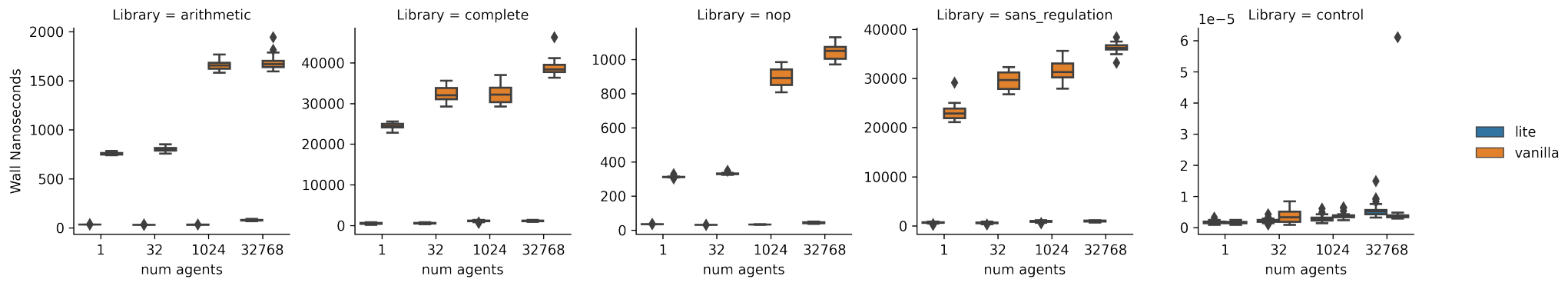

**Figure 6:** Wall time benchmarking results of 20 replicates. The x-axis represents different agent counts. Supplementary Table 1 shows raw benchmark data.

**Table 1:** Raw benchmark timings, also available as a CSV file in the supplement repository.

| Library | Implementation | Wall Nanoseconds | CPU Nanoseconds | num agents |
|---|---|---|---|---|
| arithmetic | vanilla | 760.19 | 749.88 | 1 |
| arithmetic | vanilla | 768.35 | 754.18 | 1 |
| arithmetic | vanilla | 768.87 | 755.16 | 1 |
| arithmetic | vanilla | 757.27 | 748.97 | 1 |
| arithmetic | vanilla | 759.50 | 746.07 | 1 |
| arithmetic | vanilla | 759.35 | 747.11 | 1 |
| arithmetic | vanilla | 759.50 | 746.07 | 1 |
| arithmetic | vanilla | 752.69 | 751.92 | 1 |
| arithmetic | vanilla | 741.52 | 741.52 | 1 |
| arithmetic | vanilla | 745.24 | 745.22 | 1 |
| arithmetic | vanilla | 740.75 | 740.74 | 1 |
| arithmetic | vanilla | 747.54 | 747.54 | 1 |
| arithmetic | vanilla | 745.73 | 745.74 | 1 |
| arithmetic | vanilla | 742.21 | 742.22 | 1 |
| arithmetic | vanilla | 741.18 | 741.17 | 1 |

| Library | Implementation | Wall Nanoseconds | CPU Nanoseconds | num agents |
|---|---|---|---|---|
| arithmetic | vanilla | 742.07 | 742.07 | 1 |
| arithmetic | vanilla | 746.56 | 746.56 | 1 |
| arithmetic | vanilla | 738.83 | 738.82 | 1 |
| arithmetic | vanilla | 776.20 | 776.18 | 1 |
| arithmetic | vanilla | 764.06 | 764.04 | 1 |
| arithmetic | vanilla | 769.96 | 769.93 | 1 |
| arithmetic | vanilla | 760.74 | 760.74 | 1 |
| arithmetic | vanilla | 761.06 | 761.06 | 1 |
| arithmetic | vanilla | 773.91 | 773.91 | 1 |
| arithmetic | vanilla | 779.18 | 779.18 | 1 |
| arithmetic | vanilla | 743.94 | 743.94 | 1 |
| arithmetic | vanilla | 748.39 | 748.39 | 1 |
| arithmetic | vanilla | 763.53 | 763.52 | 1 |
| arithmetic | vanilla | 750.32 | 750.30 | 1 |
| arithmetic | vanilla | 741.52 | 741.51 | 1 |
| arithmetic | vanilla | 773.12 | 773.12 | 1 |
| arithmetic | vanilla | 747.61 | 747.61 | 1 |
| arithmetic | vanilla | 750.60 | 750.60 | 1 |
| arithmetic | vanilla | 758.21 | 758.20 | 1 |
| arithmetic | vanilla | 756.08 | 756.07 | 1 |
| arithmetic | vanilla | 760.37 | 760.35 | 1 |
| arithmetic | vanilla | 765.80 | 765.78 | 1 |
| arithmetic | vanilla | 760.50 | 760.50 | 1 |
| arithmetic | vanilla | 760.77 | 760.43 | 1 |
| arithmetic | vanilla | 758.77 | 758.76 | 1 |
| arithmetic | vanilla | 765.48 | 765.48 | 1 |
| arithmetic | vanilla | 784.83 | 784.83 | 1 |
| arithmetic | vanilla | 765.47 | 765.48 | 1 |
| arithmetic | vanilla | 755.47 | 755.47 | 1 |
| arithmetic | vanilla | 781.90 | 781.89 | 1 |
| arithmetic | vanilla | 753.27 | 753.24 | 1 |
| arithmetic | vanilla | 751.04 | 751.03 | 1 |
| arithmetic | vanilla | 751.33 | 751.26 | 1 |
| arithmetic | vanilla | 749.13 | 749.13 | 1 |
| arithmetic | vanilla | 751.68 | 751.68 | 1 |
| arithmetic | vanilla | 799.42 | 799.34 | 32 |
| arithmetic | vanilla | 792.84 | 792.83 | 32 |
| arithmetic | vanilla | 807.18 | 807.17 | 32 |
| arithmetic | vanilla | 780.16 | 780.15 | 32 |
| arithmetic | vanilla | 777.25 | 777.25 | 32 |
| arithmetic | vanilla | 757.24 | 757.23 | 32 |
| arithmetic | vanilla | 801.78 | 801.72 | 32 |
| arithmetic | vanilla | 801.99 | 801.99 | 32 |
| arithmetic | vanilla | 771.23 | 771.23 | 32 |
| arithmetic | vanilla | 789.31 | 789.21 | 32 |
| arithmetic | vanilla | 813.77 | 810.35 | 32 |
| arithmetic | vanilla | 790.63 | 787.39 | 32 |
| arithmetic | vanilla | 773.45 | 772.57 | 32 |
| arithmetic | vanilla | 775.79 | 775.80 | 32 |
| arithmetic | vanilla | 769.40 | 769.40 | 32 |
| arithmetic | vanilla | 797.99 | 791.57 | 32 |
| arithmetic | vanilla | 779.22 | 779.22 | 32 |
| arithmetic | vanilla | 789.42 | 789.38 | 32 |

| Library | Implementation | Wall Nanoseconds | CPU Nanoseconds | num agents |
|---|---|---|---|---|
| arithmetic | vanilla | 851.14 | 839.15 | 32 |
| arithmetic | vanilla | 822.59 | 807.25 | 32 |
| arithmetic | vanilla | 827.15 | 812.93 | 32 |
| arithmetic | vanilla | 815.08 | 798.61 | 32 |
| arithmetic | vanilla | 827.47 | 811.93 | 32 |
| arithmetic | vanilla | 810.20 | 799.03 | 32 |
| arithmetic | vanilla | 799.69 | 797.19 | 32 |
| arithmetic | vanilla | 790.41 | 790.37 | 32 |
| arithmetic | vanilla | 797.15 | 797.11 | 32 |
| arithmetic | vanilla | 813.77 | 813.63 | 32 |
| arithmetic | vanilla | 841.25 | 841.21 | 32 |
| arithmetic | vanilla | 835.91 | 832.54 | 32 |
| arithmetic | vanilla | 834.34 | 828.55 | 32 |
| arithmetic | vanilla | 822.83 | 818.04 | 32 |
| arithmetic | vanilla | 817.66 | 813.09 | 32 |
| arithmetic | vanilla | 815.65 | 810.53 | 32 |
| arithmetic | vanilla | 823.80 | 815.10 | 32 |
| arithmetic | vanilla | 787.88 | 787.65 | 32 |
| arithmetic | vanilla | 782.20 | 782.20 | 32 |
| arithmetic | vanilla | 781.27 | 781.28 | 32 |
| arithmetic | vanilla | 788.72 | 788.72 | 32 |
| arithmetic | vanilla | 810.90 | 810.91 | 32 |
| arithmetic | vanilla | 818.32 | 818.30 | 32 |
| arithmetic | vanilla | 787.87 | 787.86 | 32 |
| arithmetic | vanilla | 802.66 | 802.60 | 32 |
| arithmetic | vanilla | 809.53 | 809.54 | 32 |
| arithmetic | vanilla | 815.69 | 815.70 | 32 |
| arithmetic | vanilla | 802.44 | 802.44 | 32 |
| arithmetic | vanilla | 831.92 | 831.90 | 32 |
| arithmetic | vanilla | 815.76 | 815.24 | 32 |
| arithmetic | vanilla | 798.05 | 798.04 | 32 |
| arithmetic | vanilla | 801.61 | 801.61 | 32 |
| arithmetic | vanilla | 1609.17 | 1609.12 | 1024 |
| arithmetic | vanilla | 1690.62 | 1690.05 | 1024 |
| arithmetic | vanilla | 1611.38 | 1609.01 | 1024 |
| arithmetic | vanilla | 1671.34 | 1670.96 | 1024 |
| arithmetic | vanilla | 1737.05 | 1730.47 | 1024 |
| arithmetic | vanilla | 1766.97 | 1766.91 | 1024 |
| arithmetic | vanilla | 1588.26 | 1588.24 | 1024 |
| arithmetic | vanilla | 1682.21 | 1680.91 | 1024 |
| arithmetic | vanilla | 1643.24 | 1643.23 | 1024 |
| arithmetic | vanilla | 1703.66 | 1703.64 | 1024 |
| arithmetic | vanilla | 1665.91 | 1665.91 | 1024 |
| arithmetic | vanilla | 1658.93 | 1653.53 | 1024 |
| arithmetic | vanilla | 1623.61 | 1616.73 | 1024 |
| arithmetic | vanilla | 1653.10 | 1653.05 | 1024 |
| arithmetic | vanilla | 1644.47 | 1644.47 | 1024 |
| arithmetic | vanilla | 1682.15 | 1682.15 | 1024 |
| arithmetic | vanilla | 1719.64 | 1719.64 | 1024 |
| arithmetic | vanilla | 1681.78 | 1681.57 | 1024 |
| arithmetic | vanilla | 1613.64 | 1613.57 | 1024 |
| arithmetic | vanilla | 1609.36 | 1608.98 | 1024 |
| arithmetic | vanilla | 1656.75 | 1656.65 | 1024 |

| Library | Implementation | Wall Nanoseconds | CPU Nanoseconds | num agents |
|---|---|---|---|---|
| arithmetic | vanilla | 1681.59 | 1681.59 | 1024 |
| arithmetic | vanilla | 1710.99 | 1710.96 | 1024 |
| arithmetic | vanilla | 1669.02 | 1669.02 | 1024 |
| arithmetic | vanilla | 1638.71 | 1638.70 | 1024 |
| arithmetic | vanilla | 1691.14 | 1691.14 | 1024 |
| arithmetic | vanilla | 1728.32 | 1728.32 | 1024 |
| arithmetic | vanilla | 1661.15 | 1661.14 | 1024 |
| arithmetic | vanilla | 1622.40 | 1622.31 | 1024 |
| arithmetic | vanilla | 1648.83 | 1648.83 | 1024 |
| arithmetic | vanilla | 1627.84 | 1625.57 | 1024 |
| arithmetic | vanilla | 1707.33 | 1707.30 | 1024 |
| arithmetic | vanilla | 1660.42 | 1658.64 | 1024 |
| arithmetic | vanilla | 1616.96 | 1616.91 | 1024 |
| arithmetic | vanilla | 1633.51 | 1633.49 | 1024 |
| arithmetic | vanilla | 1594.59 | 1594.57 | 1024 |
| arithmetic | vanilla | 1615.70 | 1615.70 | 1024 |
| arithmetic | vanilla | 1668.26 | 1668.26 | 1024 |
| arithmetic | vanilla | 1659.44 | 1659.43 | 1024 |
| arithmetic | vanilla | 1609.23 | 1609.21 | 1024 |
| arithmetic | vanilla | 1671.21 | 1670.36 | 1024 |
| arithmetic | vanilla | 1625.10 | 1625.09 | 1024 |
| arithmetic | vanilla | 1631.90 | 1631.87 | 1024 |
| arithmetic | vanilla | 1608.69 | 1608.69 | 1024 |
| arithmetic | vanilla | 1622.32 | 1617.40 | 1024 |
| arithmetic | vanilla | 1632.61 | 1627.07 | 1024 |
| arithmetic | vanilla | 1580.68 | 1580.00 | 1024 |
| arithmetic | vanilla | 1686.28 | 1686.28 | 1024 |
| arithmetic | vanilla | 1673.45 | 1646.51 | 1024 |
| arithmetic | vanilla | 1650.37 | 1613.36 | 1024 |
| arithmetic | vanilla | 1787.28 | 1778.42 | 32768 |
| arithmetic | vanilla | 1745.06 | 1712.71 | 32768 |
| arithmetic | vanilla | 1814.96 | 1782.09 | 32768 |
| arithmetic | vanilla | 1940.58 | 1939.85 | 32768 |
| arithmetic | vanilla | 1669.27 | 1669.23 | 32768 |
| arithmetic | vanilla | 1701.11 | 1700.71 | 32768 |
| arithmetic | vanilla | 1619.67 | 1619.66 | 32768 |
| arithmetic | vanilla | 1601.34 | 1601.34 | 32768 |
| arithmetic | vanilla | 1615.92 | 1615.92 | 32768 |
| arithmetic | vanilla | 1704.98 | 1704.84 | 32768 |
| arithmetic | vanilla | 1679.23 | 1679.16 | 32768 |
| arithmetic | vanilla | 1642.93 | 1642.89 | 32768 |
| arithmetic | vanilla | 1614.52 | 1614.52 | 32768 |
| arithmetic | vanilla | 1627.87 | 1627.83 | 32768 |
| arithmetic | vanilla | 1679.48 | 1679.47 | 32768 |
| arithmetic | vanilla | 1626.59 | 1626.58 | 32768 |
| arithmetic | vanilla | 1687.02 | 1686.90 | 32768 |
| arithmetic | vanilla | 1679.43 | 1679.43 | 32768 |
| arithmetic | vanilla | 1640.85 | 1640.81 | 32768 |
| arithmetic | vanilla | 1729.68 | 1728.72 | 32768 |
| arithmetic | vanilla | 1714.43 | 1714.40 | 32768 |
| arithmetic | vanilla | 1621.67 | 1621.66 | 32768 |
| arithmetic | vanilla | 1595.25 | 1595.25 | 32768 |
| arithmetic | vanilla | 1631.16 | 1631.16 | 32768 |

| Library | Implementation | Wall Nanoseconds | CPU Nanoseconds | num agents |
|---|---|---|---|---|
| arithmetic | vanilla | 1652.02 | 1651.99 | 32768 |
| arithmetic | vanilla | 1617.54 | 1617.50 | 32768 |
| arithmetic | vanilla | 1704.85 | 1702.74 | 32768 |
| arithmetic | vanilla | 1703.76 | 1693.17 | 32768 |
| arithmetic | vanilla | 1754.77 | 1754.34 | 32768 |
| arithmetic | vanilla | 1678.26 | 1639.86 | 32768 |
| arithmetic | vanilla | 1662.27 | 1634.36 | 32768 |
| arithmetic | vanilla | 1655.22 | 1629.15 | 32768 |
| arithmetic | vanilla | 1665.78 | 1636.26 | 32768 |
| arithmetic | vanilla | 1680.62 | 1648.38 | 32768 |
| arithmetic | vanilla | 1752.40 | 1728.71 | 32768 |
| arithmetic | vanilla | 1669.39 | 1669.34 | 32768 |
| arithmetic | vanilla | 1676.24 | 1672.00 | 32768 |
| arithmetic | vanilla | 1634.84 | 1634.81 | 32768 |
| arithmetic | vanilla | 1687.72 | 1687.67 | 32768 |
| arithmetic | vanilla | 1650.51 | 1650.42 | 32768 |
| arithmetic | vanilla | 1668.15 | 1668.15 | 32768 |
| arithmetic | vanilla | 1658.72 | 1658.60 | 32768 |
| arithmetic | vanilla | 1647.36 | 1647.37 | 32768 |
| arithmetic | vanilla | 1638.67 | 1638.59 | 32768 |
| arithmetic | vanilla | 1688.80 | 1688.80 | 32768 |
| arithmetic | vanilla | 1634.55 | 1634.56 | 32768 |
| arithmetic | vanilla | 1727.82 | 1727.69 | 32768 |
| arithmetic | vanilla | 1636.13 | 1636.13 | 32768 |
| arithmetic | vanilla | 1661.66 | 1661.57 | 32768 |
| arithmetic | vanilla | 1712.89 | 1691.95 | 32768 |
| complete | vanilla | 22792.39 | 22759.79 | 1 |
| complete | vanilla | 23203.62 | 23118.27 | 1 |
| complete | vanilla | 24437.95 | 24188.15 | 1 |
| complete | vanilla | 24410.70 | 24330.48 | 1 |
| complete | vanilla | 24124.18 | 24020.80 | 1 |
| complete | vanilla | 24155.33 | 23968.30 | 1 |
| complete | vanilla | 24780.18 | 24444.83 | 1 |
| complete | vanilla | 25548.30 | 25401.63 | 1 |
| complete | vanilla | 23486.37 | 23340.78 | 1 |
| complete | vanilla | 23760.35 | 23582.12 | 1 |
| complete | vanilla | 24136.77 | 23936.80 | 1 |
| complete | vanilla | 25552.98 | 25415.72 | 1 |
| complete | vanilla | 25193.67 | 24936.47 | 1 |
| complete | vanilla | 24901.96 | 24608.72 | 1 |
| complete | vanilla | 23956.94 | 23732.90 | 1 |
| complete | vanilla | 25029.21 | 24858.61 | 1 |
| complete | vanilla | 24957.04 | 24856.43 | 1 |
| complete | vanilla | 24399.10 | 24123.83 | 1 |
| complete | vanilla | 24196.05 | 23899.29 | 1 |
| complete | vanilla | 25542.33 | 25512.70 | 1 |
| complete | vanilla | 35583.70 | 35111.39 | 32 |
| complete | vanilla | 31072.12 | 30814.79 | 32 |
| complete | vanilla | 34756.04 | 34397.74 | 32 |
| complete | vanilla | 32993.73 | 32660.28 | 32 |
| complete | vanilla | 29274.59 | 29038.10 | 32 |
| complete | vanilla | 30905.56 | 30728.66 | 32 |
| complete | vanilla | 31759.00 | 31520.90 | 32 |

| Library | Implementation | Wall Nanoseconds | CPU Nanoseconds | num agents |
|---|---|---|---|---|
| complete | vanilla | 33703.31 | 33402.32 | 32 |
| complete | vanilla | 30585.90 | 30258.02 | 32 |
| complete | vanilla | 34651.71 | 34358.01 | 32 |
| complete | vanilla | 33578.90 | 33394.27 | 32 |
| complete | vanilla | 32262.49 | 32103.10 | 32 |
| complete | vanilla | 31068.13 | 30884.80 | 32 |
| complete | vanilla | 35098.88 | 34734.12 | 32 |
| complete | vanilla | 31326.04 | 31265.81 | 32 |
| complete | vanilla | 29679.55 | 29667.76 | 32 |
| complete | vanilla | 34160.38 | 33780.28 | 32 |
| complete | vanilla | 32381.72 | 32345.97 | 32 |
| complete | vanilla | 31777.50 | 31541.14 | 32 |
| complete | vanilla | 31529.37 | 31491.00 | 32 |
| complete | vanilla | 30595.67 | 30589.46 | 1024 |
| complete | vanilla | 29877.09 | 29468.47 | 1024 |
| complete | vanilla | 34749.25 | 34318.38 | 1024 |
| complete | vanilla | 32068.20 | 31617.81 | 1024 |
| complete | vanilla | 30824.03 | 30821.11 | 1024 |
| complete | vanilla | 30170.31 | 30089.37 | 1024 |
| complete | vanilla | 29264.82 | 29198.94 | 1024 |
| complete | vanilla | 30249.13 | 30214.93 | 1024 |
| complete | vanilla | 30166.86 | 30166.83 | 1024 |
| complete | vanilla | 31835.44 | 31834.66 | 1024 |
| complete | vanilla | 33134.30 | 33134.29 | 1024 |
| complete | vanilla | 32381.86 | 32343.31 | 1024 |
| complete | vanilla | 30340.41 | 30079.61 | 1024 |
| complete | vanilla | 32355.06 | 32126.01 | 1024 |
| complete | vanilla | 34694.44 | 34191.56 | 1024 |
| complete | vanilla | 36885.53 | 36467.55 | 1024 |
| complete | vanilla | 35027.90 | 34789.83 | 1024 |
| complete | vanilla | 33587.63 | 33326.76 | 1024 |
| complete | vanilla | 37003.57 | 36587.17 | 1024 |
| complete | vanilla | 32903.23 | 32439.15 | 1024 |
| complete | vanilla | 40668.84 | 40668.25 | 32768 |
| complete | vanilla | 37402.74 | 37402.17 | 32768 |
| complete | vanilla | 46248.95 | 46196.18 | 32768 |
| complete | vanilla | 40634.62 | 39885.51 | 32768 |
| complete | vanilla | 39601.10 | 38905.12 | 32768 |
| complete | vanilla | 38538.96 | 37783.22 | 32768 |
| complete | vanilla | 38794.17 | 38218.78 | 32768 |
| complete | vanilla | 38254.81 | 37851.93 | 32768 |
| complete | vanilla | 36349.80 | 36349.75 | 32768 |
| complete | vanilla | 37353.11 | 37353.03 | 32768 |
| complete | vanilla | 37442.02 | 36854.76 | 32768 |
| complete | vanilla | 37787.89 | 37014.78 | 32768 |
| complete | vanilla | 38519.35 | 37777.28 | 32768 |
| complete | vanilla | 37983.14 | 37299.06 | 32768 |
| complete | vanilla | 39508.65 | 38880.46 | 32768 |
| complete | vanilla | 38163.34 | 37575.59 | 32768 |
| complete | vanilla | 37972.43 | 37299.13 | 32768 |
| complete | vanilla | 41130.09 | 40588.81 | 32768 |
| complete | vanilla | 38845.03 | 38391.07 | 32768 |
| complete | vanilla | 37005.04 | 37005.05 | 32768 |

| Library | Implementation | Wall Nanoseconds | CPU Nanoseconds | num agents |
|---|---|---|---|---|
| nop | vanilla | 303.52 | 301.90 | 1 |
| nop | vanilla | 328.01 | 327.21 | 1 |
| nop | vanilla | 314.50 | 312.63 | 1 |
| nop | vanilla | 313.99 | 309.94 | 1 |
| nop | vanilla | 310.24 | 306.25 | 1 |
| nop | vanilla | 312.32 | 308.06 | 1 |
| nop | vanilla | 309.92 | 305.78 | 1 |
| nop | vanilla | 311.39 | 307.72 | 1 |
| nop | vanilla | 311.69 | 307.99 | 1 |
| nop | vanilla | 310.43 | 305.72 | 1 |
| nop | vanilla | 310.99 | 305.76 | 1 |
| nop | vanilla | 310.44 | 306.11 | 1 |
| nop | vanilla | 308.91 | 304.43 | 1 |
| nop | vanilla | 318.68 | 314.04 | 1 |
| nop | vanilla | 312.30 | 307.24 | 1 |
| nop | vanilla | 310.81 | 307.19 | 1 |
| nop | vanilla | 312.34 | 307.74 | 1 |
| nop | vanilla | 313.65 | 308.08 | 1 |
| nop | vanilla | 310.85 | 305.64 | 1 |
| nop | vanilla | 309.23 | 305.25 | 1 |
| nop | vanilla | 330.75 | 326.61 | 32 |
| nop | vanilla | 332.66 | 328.22 | 32 |
| nop | vanilla | 332.90 | 327.83 | 32 |
| nop | vanilla | 333.40 | 329.06 | 32 |
| nop | vanilla | 333.46 | 327.44 | 32 |
| nop | vanilla | 334.44 | 328.47 | 32 |
| nop | vanilla | 332.10 | 329.81 | 32 |
| nop | vanilla | 326.89 | 326.89 | 32 |
| nop | vanilla | 325.05 | 325.05 | 32 |
| nop | vanilla | 327.66 | 327.66 | 32 |
| nop | vanilla | 328.53 | 327.30 | 32 |
| nop | vanilla | 328.30 | 328.30 | 32 |
| nop | vanilla | 328.44 | 328.44 | 32 |
| nop | vanilla | 329.51 | 329.51 | 32 |
| nop | vanilla | 324.31 | 324.30 | 32 |
| nop | vanilla | 330.28 | 330.27 | 32 |
| nop | vanilla | 330.20 | 330.20 | 32 |
| nop | vanilla | 326.90 | 326.89 | 32 |
| nop | vanilla | 336.34 | 336.34 | 32 |
| nop | vanilla | 348.24 | 348.21 | 32 |
| nop | vanilla | 961.18 | 961.13 | 1024 |
| nop | vanilla | 932.15 | 932.13 | 1024 |
| nop | vanilla | 985.26 | 985.26 | 1024 |
| nop | vanilla | 957.85 | 957.84 | 1024 |
| nop | vanilla | 942.41 | 928.07 | 1024 |
| nop | vanilla | 906.35 | 890.08 | 1024 |
| nop | vanilla | 888.76 | 884.23 | 1024 |
| nop | vanilla | 940.81 | 940.79 | 1024 |
| nop | vanilla | 859.19 | 859.19 | 1024 |
| nop | vanilla | 837.66 | 837.66 | 1024 |
| nop | vanilla | 848.95 | 848.95 | 1024 |
| nop | vanilla | 845.00 | 844.98 | 1024 |
| nop | vanilla | 895.59 | 895.57 | 1024 |

| Library | Implementation | Wall Nanoseconds | CPU Nanoseconds | num agents |
|---|---|---|---|---|
| nop | vanilla | 851.09 | 851.07 | 1024 |
| nop | vanilla | 858.08 | 858.08 | 1024 |
| nop | vanilla | 849.78 | 849.78 | 1024 |
| nop | vanilla | 867.20 | 867.20 | 1024 |
| nop | vanilla | 904.93 | 904.90 | 1024 |
| nop | vanilla | 807.05 | 806.04 | 1024 |
| nop | vanilla | 971.06 | 967.02 | 1024 |
| nop | vanilla | 1112.17 | 1111.97 | 32768 |
| nop | vanilla | 1082.29 | 1082.27 | 32768 |
| nop | vanilla | 1107.78 | 1107.65 | 32768 |
| nop | vanilla | 1056.07 | 1056.07 | 32768 |
| nop | vanilla | 1070.70 | 1070.69 | 32768 |
| nop | vanilla | 1070.97 | 1070.95 | 32768 |
| nop | vanilla | 1034.78 | 1034.74 | 32768 |
| nop | vanilla | 1062.35 | 1062.35 | 32768 |
| nop | vanilla | 1001.25 | 997.63 | 32768 |
| nop | vanilla | 1004.38 | 1004.36 | 32768 |
| nop | vanilla | 1042.76 | 1042.76 | 32768 |
| nop | vanilla | 973.33 | 973.20 | 32768 |
| nop | vanilla | 1021.51 | 1021.51 | 32768 |
| nop | vanilla | 1004.24 | 1004.24 | 32768 |
| nop | vanilla | 1129.32 | 1129.27 | 32768 |
| nop | vanilla | 1071.88 | 1071.88 | 32768 |
| nop | vanilla | 1044.98 | 1044.98 | 32768 |
| nop | vanilla | 1075.35 | 1075.33 | 32768 |
| nop | vanilla | 998.79 | 998.76 | 32768 |
| nop | vanilla | 970.16 | 970.16 | 32768 |
| sans_regulation | vanilla | 23738.83 | 23687.11 | 1 |
| sans_regulation | vanilla | 29138.34 | 28867.22 | 1 |
| sans_regulation | vanilla | 23012.36 | 23011.81 | 1 |
| sans_regulation | vanilla | 22879.23 | 22879.28 | 1 |
| sans_regulation | vanilla | 21606.44 | 21606.06 | 1 |
| sans_regulation | vanilla | 24315.65 | 24189.47 | 1 |
| sans_regulation | vanilla | 24290.39 | 24256.59 | 1 |
| sans_regulation | vanilla | 22075.37 | 22074.20 | 1 |
| sans_regulation | vanilla | 24206.39 | 24205.97 | 1 |
| sans_regulation | vanilla | 23650.88 | 23650.94 | 1 |
| sans_regulation | vanilla | 21503.51 | 21503.29 | 1 |
| sans_regulation | vanilla | 22603.93 | 22556.04 | 1 |
| sans_regulation | vanilla | 22832.25 | 22832.31 | 1 |
| sans_regulation | vanilla | 21128.39 | 21128.38 | 1 |
| sans_regulation | vanilla | 21393.67 | 21393.72 | 1 |
| sans_regulation | vanilla | 23769.54 | 23769.58 | 1 |
| sans_regulation | vanilla | 21918.97 | 21918.47 | 1 |
| sans_regulation | vanilla | 21961.28 | 21960.96 | 1 |
| sans_regulation | vanilla | 24999.87 | 24999.54 | 1 |
| sans_regulation | vanilla | 22841.80 | 22807.61 | 1 |
| sans_regulation | vanilla | 30745.11 | 30744.65 | 32 |
| sans_regulation | vanilla | 30997.85 | 30997.52 | 32 |
| sans_regulation | vanilla | 26773.27 | 26772.85 | 32 |
| sans_regulation | vanilla | 26852.47 | 26780.96 | 32 |
| sans_regulation | vanilla | 28659.17 | 28658.56 | 32 |
| sans_regulation | vanilla | 28455.67 | 28433.84 | 32 |

| Library | Implementation | Wall Nanoseconds | CPU Nanoseconds | num agents |
|---|---|---|---|---|
| sans_regulation | vanilla | 30140.91 | 30125.24 | 32 |
| sans_regulation | vanilla | 31165.90 | 31114.38 | 32 |
| sans_regulation | vanilla | 31729.55 | 31530.98 | 32 |
| sans_regulation | vanilla | 27850.57 | 27517.93 | 32 |
| sans_regulation | vanilla | 27863.73 | 27490.69 | 32 |
| sans_regulation | vanilla | 29210.21 | 28824.14 | 32 |
| sans_regulation | vanilla | 31009.49 | 30626.25 | 32 |
| sans_regulation | vanilla | 31467.94 | 30930.42 | 32 |
| sans_regulation | vanilla | 31278.67 | 31062.19 | 32 |
| sans_regulation | vanilla | 27853.75 | 27853.73 | 32 |
| sans_regulation | vanilla | 27778.20 | 27775.60 | 32 |
| sans_regulation | vanilla | 32296.03 | 32295.63 | 32 |
| sans_regulation | vanilla | 31422.35 | 31360.82 | 32 |
| sans_regulation | vanilla | 28538.44 | 28536.98 | 32 |
| sans_regulation | vanilla | 30988.46 | 30987.94 | 1024 |
| sans_regulation | vanilla | 30206.44 | 30206.39 | 1024 |
| sans_regulation | vanilla | 31261.62 | 31258.95 | 1024 |
| sans_regulation | vanilla | 30122.97 | 30122.03 | 1024 |
| sans_regulation | vanilla | 35259.54 | 35259.21 | 1024 |
| sans_regulation | vanilla | 30799.63 | 30796.55 | 1024 |
| sans_regulation | vanilla | 29496.77 | 29495.44 | 1024 |
| sans_regulation | vanilla | 31268.15 | 31266.79 | 1024 |
| sans_regulation | vanilla | 32659.25 | 32658.48 | 1024 |
| sans_regulation | vanilla | 29186.15 | 29185.87 | 1024 |
| sans_regulation | vanilla | 31275.83 | 31275.84 | 1024 |
| sans_regulation | vanilla | 27903.06 | 27902.45 | 1024 |
| sans_regulation | vanilla | 32888.62 | 32887.92 | 1024 |
| sans_regulation | vanilla | 29790.59 | 29788.20 | 1024 |
| sans_regulation | vanilla | 33735.68 | 33735.26 | 1024 |
| sans_regulation | vanilla | 35563.13 | 35560.50 | 1024 |
| sans_regulation | vanilla | 30951.49 | 30758.96 | 1024 |
| sans_regulation | vanilla | 34846.64 | 34189.19 | 1024 |
| sans_regulation | vanilla | 33492.69 | 32931.72 | 1024 |
| sans_regulation | vanilla | 32136.12 | 31705.89 | 1024 |
| sans_regulation | vanilla | 36815.20 | 36814.18 | 32768 |
| sans_regulation | vanilla | 37040.28 | 37040.17 | 32768 |
| sans_regulation | vanilla | 36282.99 | 36282.96 | 32768 |
| sans_regulation | vanilla | 38338.96 | 38337.87 | 32768 |
| sans_regulation | vanilla | 35984.65 | 35983.68 | 32768 |
| sans_regulation | vanilla | 35774.73 | 35774.66 | 32768 |
| sans_regulation | vanilla | 36623.65 | 36623.81 | 32768 |
| sans_regulation | vanilla | 34905.90 | 34904.35 | 32768 |
| sans_regulation | vanilla | 33155.05 | 33155.13 | 32768 |
| sans_regulation | vanilla | 36185.56 | 36184.49 | 32768 |
| sans_regulation | vanilla | 35980.15 | 35968.70 | 32768 |
| sans_regulation | vanilla | 35395.24 | 35395.27 | 32768 |
| sans_regulation | vanilla | 35854.08 | 35669.65 | 32768 |
| sans_regulation | vanilla | 37071.69 | 36384.46 | 32768 |
| sans_regulation | vanilla | 37510.02 | 36879.27 | 32768 |
| sans_regulation | vanilla | 36601.63 | 36601.42 | 32768 |
| sans_regulation | vanilla | 36582.21 | 36580.73 | 32768 |
| sans_regulation | vanilla | 36695.07 | 36695.07 | 32768 |
| sans_regulation | vanilla | 35868.59 | 35867.27 | 32768 |

| Library | Implementation | Wall Nanoseconds | CPU Nanoseconds | num agents |
|---|---|---|---|---|
| sans_regulation | vanilla | 35831.21 | 35830.75 | 32768 |
| control | vanilla | 0.00 | 0.00 | 1 |
| control | vanilla | 0.00 | 0.00 | 1 |
| control | vanilla | 0.00 | 0.00 | 1 |
| control | vanilla | 0.00 | 0.00 | 1 |
| control | vanilla | 0.00 | 0.00 | 1 |
| control | vanilla | 0.00 | 0.00 | 1 |
| control | vanilla | 0.00 | 0.00 | 1 |
| control | vanilla | 0.00 | 0.00 | 1 |
| control | vanilla | 0.00 | 0.00 | 1 |
| control | vanilla | 0.00 | 0.00 | 1 |
| control | vanilla | 0.00 | 0.00 | 1 |
| control | vanilla | 0.00 | 0.00 | 1 |
| control | vanilla | 0.00 | 0.00 | 1 |
| control | vanilla | 0.00 | 0.00 | 1 |
| control | vanilla | 0.00 | 0.00 | 1 |
| control | vanilla | 0.00 | 0.00 | 1 |
| control | vanilla | 0.00 | 0.00 | 1 |
| control | vanilla | 0.00 | 0.00 | 1 |
| control | vanilla | 0.00 | 0.00 | 1 |
| control | vanilla | 0.00 | 0.00 | 1 |
| control | vanilla | 0.00 | 0.00 | 32 |
| control | vanilla | 0.00 | 0.00 | 32 |
| control | vanilla | 0.00 | 0.00 | 32 |
| control | vanilla | 0.00 | 0.00 | 32 |
| control | vanilla | 0.00 | 0.00 | 32 |
| control | vanilla | 0.00 | 0.00 | 32 |
| control | vanilla | 0.00 | 0.00 | 32 |
| control | vanilla | 0.00 | 0.00 | 32 |
| control | vanilla | 0.00 | 0.00 | 32 |
| control | vanilla | 0.00 | 0.00 | 32 |
| control | vanilla | 0.00 | 0.00 | 32 |
| control | vanilla | 0.00 | 0.00 | 32 |
| control | vanilla | 0.00 | 0.00 | 32 |
| control | vanilla | 0.00 | 0.00 | 32 |
| control | vanilla | 0.00 | 0.00 | 32 |
| control | vanilla | 0.00 | 0.00 | 32 |
| control | vanilla | 0.00 | 0.00 | 32 |
| control | vanilla | 0.00 | 0.00 | 32 |
| control | vanilla | 0.00 | 0.00 | 32 |
| control | vanilla | 0.00 | 0.00 | 32 |
| control | vanilla | 0.00 | 0.00 | 1024 |
| control | vanilla | 0.00 | 0.00 | 1024 |
| control | vanilla | 0.00 | 0.00 | 1024 |
| control | vanilla | 0.00 | 0.00 | 1024 |
| control | vanilla | 0.00 | 0.00 | 1024 |
| control | vanilla | 0.00 | 0.00 | 1024 |
| control | vanilla | 0.00 | 0.00 | 1024 |
| control | vanilla | 0.00 | 0.00 | 1024 |
| control | vanilla | 0.00 | 0.00 | 1024 |
| control | vanilla | 0.00 | 0.00 | 1024 |
| control | vanilla | 0.00 | 0.00 | 1024 |
| control | vanilla | 0.00 | 0.00 | 1024 |

| Library | Implementation | Wall Nanoseconds | CPU Nanoseconds | num agents |
|---|---|---|---|---|
| control | vanilla | 0.00 | 0.00 | 1024 |
| control | vanilla | 0.00 | 0.00 | 1024 |
| control | vanilla | 0.00 | 0.00 | 1024 |
| control | vanilla | 0.00 | 0.00 | 1024 |
| control | vanilla | 0.00 | 0.00 | 1024 |
| control | vanilla | 0.00 | 0.00 | 1024 |
| control | vanilla | 0.00 | 0.00 | 1024 |
| control | vanilla | 0.00 | 0.00 | 1024 |
| control | vanilla | 0.00 | 0.00 | 32768 |
| control | vanilla | 0.00 | 0.00 | 32768 |
| control | vanilla | 0.00 | 0.00 | 32768 |
| control | vanilla | 0.00 | 0.00 | 32768 |
| control | vanilla | 0.00 | 0.00 | 32768 |
| control | vanilla | 0.00 | 0.00 | 32768 |
| control | vanilla | 0.00 | 0.00 | 32768 |
| control | vanilla | 0.00 | 0.00 | 32768 |
| control | vanilla | 0.00 | 0.00 | 32768 |
| control | vanilla | 0.00 | 0.00 | 32768 |
| control | vanilla | 0.00 | 0.00 | 32768 |
| control | vanilla | 0.00 | 0.00 | 32768 |
| control | vanilla | 0.00 | 0.00 | 32768 |
| control | vanilla | 0.00 | 0.00 | 32768 |
| control | vanilla | 0.00 | 0.00 | 32768 |
| control | vanilla | 0.00 | 0.00 | 32768 |
| control | vanilla | 0.00 | 0.00 | 32768 |
| control | vanilla | 0.00 | 0.00 | 32768 |
| control | vanilla | 0.00 | 0.00 | 32768 |
| arithmetic | lite | 37.26 | 36.94 | 1 |
| arithmetic | lite | 36.40 | 36.40 | 1 |
| arithmetic | lite | 36.07 | 36.07 | 1 |
| arithmetic | lite | 35.99 | 35.99 | 1 |
| arithmetic | lite | 36.26 | 36.26 | 1 |
| arithmetic | lite | 36.76 | 36.73 | 1 |
| arithmetic | lite | 36.77 | 36.77 | 1 |
| arithmetic | lite | 36.34 | 36.34 | 1 |
| arithmetic | lite | 36.18 | 36.18 | 1 |
| arithmetic | lite | 36.05 | 36.05 | 1 |
| arithmetic | lite | 36.25 | 36.25 | 1 |
| arithmetic | lite | 36.48 | 36.48 | 1 |
| arithmetic | lite | 35.84 | 35.84 | 1 |
| arithmetic | lite | 36.44 | 36.44 | 1 |
| arithmetic | lite | 36.25 | 36.25 | 1 |
| arithmetic | lite | 36.15 | 36.15 | 1 |
| arithmetic | lite | 36.12 | 36.12 | 1 |
| arithmetic | lite | 36.32 | 36.32 | 1 |
| arithmetic | lite | 36.30 | 36.30 | 1 |
| arithmetic | lite | 36.05 | 36.05 | 1 |
| arithmetic | lite | 36.00 | 36.00 | 1 |
| arithmetic | lite | 35.95 | 35.95 | 1 |
| arithmetic | lite | 35.98 | 35.98 | 1 |
| arithmetic | lite | 36.10 | 36.10 | 1 |
| arithmetic | lite | 36.29 | 36.29 | 1 |

| Library | Implementation | Wall Nanoseconds | CPU Nanoseconds | num agents |
|---|---|---|---|---|
| arithmetic | lite | 36.18 | 36.18 | 1 |
| arithmetic | lite | 36.03 | 36.03 | 1 |
| arithmetic | lite | 35.97 | 35.96 | 1 |
| arithmetic | lite | 36.28 | 36.28 | 1 |
| arithmetic | lite | 36.18 | 36.18 | 1 |
| arithmetic | lite | 35.90 | 35.90 | 1 |
| arithmetic | lite | 36.09 | 36.09 | 1 |
| arithmetic | lite | 36.35 | 36.35 | 1 |
| arithmetic | lite | 35.98 | 35.98 | 1 |
| arithmetic | lite | 35.88 | 35.88 | 1 |
| arithmetic | lite | 35.90 | 35.90 | 1 |
| arithmetic | lite | 36.32 | 36.32 | 1 |
| arithmetic | lite | 36.97 | 36.62 | 1 |
| arithmetic | lite | 37.05 | 36.63 | 1 |
| arithmetic | lite | 37.25 | 36.58 | 1 |
| arithmetic | lite | 36.59 | 36.10 | 1 |
| arithmetic | lite | 36.75 | 36.23 | 1 |
| arithmetic | lite | 36.32 | 36.18 | 1 |
| arithmetic | lite | 35.98 | 35.98 | 1 |
| arithmetic | lite | 36.02 | 36.02 | 1 |
| arithmetic | lite | 36.06 | 36.06 | 1 |
| arithmetic | lite | 35.94 | 35.94 | 1 |
| arithmetic | lite | 36.60 | 36.60 | 1 |
| arithmetic | lite | 35.73 | 35.73 | 1 |
| arithmetic | lite | 36.41 | 36.41 | 1 |
| arithmetic | lite | 32.42 | 32.42 | 32 |
| arithmetic | lite | 32.56 | 32.56 | 32 |
| arithmetic | lite | 33.26 | 33.26 | 32 |
| arithmetic | lite | 32.74 | 32.74 | 32 |
| arithmetic | lite | 32.57 | 32.57 | 32 |
| arithmetic | lite | 32.49 | 32.49 | 32 |
| arithmetic | lite | 32.51 | 32.51 | 32 |
| arithmetic | lite | 32.73 | 32.73 | 32 |
| arithmetic | lite | 32.68 | 32.68 | 32 |
| arithmetic | lite | 32.69 | 32.69 | 32 |
| arithmetic | lite | 32.73 | 32.73 | 32 |
| arithmetic | lite | 32.69 | 32.69 | 32 |
| arithmetic | lite | 32.65 | 32.65 | 32 |
| arithmetic | lite | 32.78 | 32.78 | 32 |
| arithmetic | lite | 33.58 | 33.58 | 32 |
| arithmetic | lite | 33.83 | 33.83 | 32 |
| arithmetic | lite | 33.18 | 33.18 | 32 |
| arithmetic | lite | 32.96 | 32.96 | 32 |
| arithmetic | lite | 34.01 | 34.01 | 32 |
| arithmetic | lite | 32.99 | 32.99 | 32 |
| arithmetic | lite | 32.74 | 32.74 | 32 |
| arithmetic | lite | 33.47 | 33.47 | 32 |
| arithmetic | lite | 33.22 | 33.22 | 32 |
| arithmetic | lite | 32.39 | 32.39 | 32 |
| arithmetic | lite | 32.52 | 32.52 | 32 |
| arithmetic | lite | 32.57 | 32.57 | 32 |
| arithmetic | lite | 32.56 | 32.56 | 32 |
| arithmetic | lite | 32.48 | 32.48 | 32 |

| Library | Implementation | Wall Nanoseconds | CPU Nanoseconds | num agents |
|---|---|---|---|---|
| arithmetic | lite | 32.74 | 32.74 | 32 |
| arithmetic | lite | 32.64 | 32.64 | 32 |
| arithmetic | lite | 32.69 | 32.69 | 32 |
| arithmetic | lite | 33.21 | 33.21 | 32 |
| arithmetic | lite | 32.98 | 32.98 | 32 |
| arithmetic | lite | 32.60 | 32.60 | 32 |
| arithmetic | lite | 32.64 | 32.64 | 32 |
| arithmetic | lite | 32.49 | 32.49 | 32 |
| arithmetic | lite | 32.53 | 32.53 | 32 |
| arithmetic | lite | 32.53 | 32.53 | 32 |
| arithmetic | lite | 32.78 | 32.78 | 32 |
| arithmetic | lite | 32.92 | 32.92 | 32 |
| arithmetic | lite | 32.49 | 32.49 | 32 |
| arithmetic | lite | 32.57 | 32.57 | 32 |
| arithmetic | lite | 32.59 | 32.59 | 32 |
| arithmetic | lite | 32.67 | 32.67 | 32 |
| arithmetic | lite | 32.78 | 32.77 | 32 |
| arithmetic | lite | 32.47 | 32.47 | 32 |
| arithmetic | lite | 32.58 | 32.57 | 32 |
| arithmetic | lite | 32.44 | 32.44 | 32 |
| arithmetic | lite | 32.58 | 32.58 | 32 |
| arithmetic | lite | 32.61 | 32.60 | 32 |
| arithmetic | lite | 32.84 | 32.84 | 1024 |
| arithmetic | lite | 32.52 | 32.52 | 1024 |
| arithmetic | lite | 33.35 | 33.35 | 1024 |
| arithmetic | lite | 32.62 | 32.62 | 1024 |
| arithmetic | lite | 32.91 | 32.91 | 1024 |
| arithmetic | lite | 32.91 | 32.91 | 1024 |
| arithmetic | lite | 33.28 | 33.28 | 1024 |
| arithmetic | lite | 33.32 | 33.32 | 1024 |
| arithmetic | lite | 33.13 | 33.13 | 1024 |
| arithmetic | lite | 34.21 | 34.21 | 1024 |
| arithmetic | lite | 33.92 | 33.92 | 1024 |
| arithmetic | lite | 34.18 | 34.18 | 1024 |
| arithmetic | lite | 34.92 | 34.92 | 1024 |
| arithmetic | lite | 34.45 | 34.45 | 1024 |
| arithmetic | lite | 33.67 | 33.67 | 1024 |
| arithmetic | lite | 33.54 | 33.54 | 1024 |
| arithmetic | lite | 33.39 | 33.39 | 1024 |
| arithmetic | lite | 33.65 | 33.65 | 1024 |
| arithmetic | lite | 33.99 | 33.99 | 1024 |
| arithmetic | lite | 33.51 | 33.51 | 1024 |
| arithmetic | lite | 33.94 | 33.94 | 1024 |
| arithmetic | lite | 34.04 | 34.04 | 1024 |
| arithmetic | lite | 32.59 | 32.59 | 1024 |
| arithmetic | lite | 33.38 | 33.38 | 1024 |
| arithmetic | lite | 32.87 | 32.87 | 1024 |
| arithmetic | lite | 33.95 | 33.95 | 1024 |
| arithmetic | lite | 33.13 | 33.13 | 1024 |
| arithmetic | lite | 33.81 | 33.81 | 1024 |
| arithmetic | lite | 33.73 | 33.73 | 1024 |
| arithmetic | lite | 33.93 | 33.93 | 1024 |
| arithmetic | lite | 33.84 | 33.84 | 1024 |

| Library | Implementation | Wall Nanoseconds | CPU Nanoseconds | num agents |
|---|---|---|---|---|
| arithmetic | lite | 33.93 | 33.93 | 1024 |
| arithmetic | lite | 34.36 | 34.36 | 1024 |
| arithmetic | lite | 33.30 | 33.30 | 1024 |
| arithmetic | lite | 33.18 | 33.18 | 1024 |
| arithmetic | lite | 33.14 | 33.14 | 1024 |
| arithmetic | lite | 33.64 | 33.64 | 1024 |
| arithmetic | lite | 33.22 | 33.22 | 1024 |
| arithmetic | lite | 33.11 | 33.11 | 1024 |
| arithmetic | lite | 32.80 | 32.79 | 1024 |
| arithmetic | lite | 32.72 | 32.72 | 1024 |
| arithmetic | lite | 33.32 | 33.31 | 1024 |
| arithmetic | lite | 33.22 | 33.22 | 1024 |
| arithmetic | lite | 32.57 | 32.57 | 1024 |
| arithmetic | lite | 32.89 | 32.89 | 1024 |
| arithmetic | lite | 33.38 | 33.38 | 1024 |
| arithmetic | lite | 33.72 | 33.72 | 1024 |
| arithmetic | lite | 33.45 | 33.44 | 1024 |
| arithmetic | lite | 33.23 | 33.23 | 1024 |
| arithmetic | lite | 33.09 | 33.09 | 1024 |
| arithmetic | lite | 74.96 | 74.96 | 32768 |
| arithmetic | lite | 77.06 | 77.05 | 32768 |
| arithmetic | lite | 71.70 | 71.70 | 32768 |
| arithmetic | lite | 74.28 | 74.28 | 32768 |
| arithmetic | lite | 79.42 | 79.42 | 32768 |
| arithmetic | lite | 89.80 | 89.80 | 32768 |
| arithmetic | lite | 83.12 | 83.12 | 32768 |
| arithmetic | lite | 82.23 | 82.23 | 32768 |
| arithmetic | lite | 84.08 | 84.08 | 32768 |
| arithmetic | lite | 76.35 | 76.35 | 32768 |
| arithmetic | lite | 82.43 | 82.43 | 32768 |
| arithmetic | lite | 92.88 | 92.65 | 32768 |
| arithmetic | lite | 80.75 | 80.23 | 32768 |
| arithmetic | lite | 83.53 | 83.24 | 32768 |
| arithmetic | lite | 87.40 | 87.35 | 32768 |
| arithmetic | lite | 84.98 | 84.98 | 32768 |
| arithmetic | lite | 85.25 | 85.23 | 32768 |
| arithmetic | lite | 80.18 | 80.18 | 32768 |
| arithmetic | lite | 74.47 | 74.47 | 32768 |
| arithmetic | lite | 74.64 | 74.51 | 32768 |
| arithmetic | lite | 73.88 | 73.83 | 32768 |
| arithmetic | lite | 80.99 | 80.99 | 32768 |
| arithmetic | lite | 80.54 | 80.53 | 32768 |
| arithmetic | lite | 71.49 | 71.49 | 32768 |
| arithmetic | lite | 74.67 | 74.66 | 32768 |
| arithmetic | lite | 73.00 | 72.80 | 32768 |
| arithmetic | lite | 78.21 | 76.67 | 32768 |
| arithmetic | lite | 80.96 | 80.55 | 32768 |
| arithmetic | lite | 78.41 | 77.95 | 32768 |
| arithmetic | lite | 81.09 | 80.90 | 32768 |
| arithmetic | lite | 84.24 | 84.24 | 32768 |
| arithmetic | lite | 80.57 | 80.57 | 32768 |
| arithmetic | lite | 82.49 | 82.49 | 32768 |
| arithmetic | lite | 87.69 | 87.55 | 32768 |

| Library | Implementation | Wall Nanoseconds | CPU Nanoseconds | num agents |
|---|---|---|---|---|
| arithmetic | lite | 78.81 | 78.47 | 32768 |
| arithmetic | lite | 74.21 | 74.18 | 32768 |
| arithmetic | lite | 73.34 | 73.32 | 32768 |
| arithmetic | lite | 76.17 | 76.16 | 32768 |
| arithmetic | lite | 79.14 | 78.89 | 32768 |
| arithmetic | lite | 70.46 | 70.46 | 32768 |
| arithmetic | lite | 70.57 | 70.54 | 32768 |
| arithmetic | lite | 72.08 | 72.08 | 32768 |
| arithmetic | lite | 78.32 | 78.32 | 32768 |
| arithmetic | lite | 70.07 | 70.07 | 32768 |
| arithmetic | lite | 68.57 | 68.57 | 32768 |
| arithmetic | lite | 76.20 | 76.20 | 32768 |
| arithmetic | lite | 79.14 | 79.13 | 32768 |
| arithmetic | lite | 86.78 | 86.59 | 32768 |
| arithmetic | lite | 73.84 | 73.82 | 32768 |
| arithmetic | lite | 80.13 | 80.11 | 32768 |
| complete | lite | 723.53 | 722.56 | 1 |
| complete | lite | 689.93 | 689.71 | 1 |
| complete | lite | 624.18 | 624.17 | 1 |
| complete | lite | 653.93 | 653.92 | 1 |
| complete | lite | 423.63 | 423.63 | 1 |
| complete | lite | 792.42 | 792.39 | 1 |
| complete | lite | 511.51 | 511.51 | 1 |
| complete | lite | 749.90 | 749.81 | 1 |
| complete | lite | 388.26 | 388.26 | 1 |
| complete | lite | 305.39 | 305.39 | 1 |
| complete | lite | 740.99 | 740.98 | 1 |
| complete | lite | 553.88 | 553.85 | 1 |
| complete | lite | 194.43 | 194.43 | 1 |
| complete | lite | 601.60 | 601.60 | 1 |
| complete | lite | 545.33 | 544.80 | 1 |
| complete | lite | 674.82 | 674.77 | 1 |
| complete | lite | 170.13 | 170.13 | 1 |
| complete | lite | 587.25 | 587.21 | 1 |
| complete | lite | 322.85 | 322.85 | 1 |
| complete | lite | 563.87 | 563.63 | 1 |
| complete | lite | 609.05 | 609.05 | 1 |
| complete | lite | 306.31 | 306.30 | 1 |
| complete | lite | 218.02 | 218.02 | 1 |
| complete | lite | 674.58 | 674.36 | 1 |
| complete | lite | 295.65 | 294.63 | 1 |
| complete | lite | 314.86 | 314.87 | 1 |
| complete | lite | 662.84 | 662.84 | 1 |
| complete | lite | 517.38 | 517.04 | 1 |
| complete | lite | 632.43 | 632.40 | 1 |
| complete | lite | 754.63 | 752.63 | 1 |
| complete | lite | 800.15 | 794.34 | 1 |
| complete | lite | 681.67 | 679.89 | 1 |
| complete | lite | 201.38 | 201.37 | 1 |
| complete | lite | 627.25 | 627.25 | 1 |
| complete | lite | 554.92 | 554.91 | 1 |
| complete | lite | 669.58 | 669.57 | 1 |
| complete | lite | 470.09 | 470.06 | 1 |

| Library | Implementation | Wall Nanoseconds | CPU Nanoseconds | num agents |
|---|---|---|---|---|
| complete | lite | 640.28 | 640.28 | 1 |
| complete | lite | 165.81 | 165.80 | 1 |
| complete | lite | 581.43 | 581.42 | 1 |
| complete | lite | 649.63 | 649.62 | 1 |
| complete | lite | 588.29 | 588.29 | 1 |
| complete | lite | 639.92 | 639.91 | 1 |
| complete | lite | 574.92 | 574.93 | 1 |
| complete | lite | 260.30 | 260.30 | 1 |
| complete | lite | 600.60 | 600.59 | 1 |
| complete | lite | 683.82 | 683.77 | 1 |
| complete | lite | 287.50 | 287.48 | 1 |
| complete | lite | 755.56 | 755.56 | 1 |
| complete | lite | 747.17 | 747.16 | 1 |
| complete | lite | 776.32 | 776.27 | 32 |
| complete | lite | 543.08 | 543.08 | 32 |
| complete | lite | 412.97 | 412.96 | 32 |
| complete | lite | 674.31 | 674.29 | 32 |
| complete | lite | 634.87 | 634.85 | 32 |
| complete | lite | 512.69 | 512.68 | 32 |
| complete | lite | 691.37 | 691.37 | 32 |
| complete | lite | 633.43 | 633.42 | 32 |
| complete | lite | 489.41 | 489.36 | 32 |
| complete | lite | 513.72 | 513.72 | 32 |
| complete | lite | 309.70 | 309.70 | 32 |
| complete | lite | 495.75 | 495.72 | 32 |
| complete | lite | 569.48 | 569.48 | 32 |
| complete | lite | 599.90 | 599.91 | 32 |
| complete | lite | 607.71 | 607.71 | 32 |
| complete | lite | 734.97 | 734.92 | 32 |
| complete | lite | 593.86 | 593.85 | 32 |
| complete | lite | 310.90 | 310.90 | 32 |
| complete | lite | 370.72 | 370.71 | 32 |
| complete | lite | 661.21 | 661.20 | 32 |
| complete | lite | 483.80 | 483.79 | 32 |
| complete | lite | 620.93 | 620.94 | 32 |
| complete | lite | 395.31 | 395.31 | 32 |
| complete | lite | 634.60 | 634.59 | 32 |
| complete | lite | 658.06 | 658.06 | 32 |
| complete | lite | 644.53 | 644.53 | 32 |
| complete | lite | 585.79 | 585.79 | 32 |
| complete | lite | 431.48 | 431.41 | 32 |
| complete | lite | 778.63 | 778.41 | 32 |
| complete | lite | 504.66 | 504.63 | 32 |
| complete | lite | 342.39 | 342.37 | 32 |
| complete | lite | 568.08 | 567.99 | 32 |
| complete | lite | 817.80 | 817.79 | 32 |
| complete | lite | 390.51 | 390.51 | 32 |
| complete | lite | 587.08 | 587.06 | 32 |
| complete | lite | 416.91 | 416.88 | 32 |
| complete | lite | 841.17 | 841.17 | 32 |
| complete | lite | 626.75 | 626.73 | 32 |
| complete | lite | 765.55 | 765.55 | 32 |
| complete | lite | 475.86 | 475.86 | 32 |

| Library | Implementation | Wall Nanoseconds | CPU Nanoseconds | num agents |
|---|---|---|---|---|
| complete | lite | 480.66 | 480.66 | 32 |
| complete | lite | 395.58 | 395.58 | 32 |
| complete | lite | 703.39 | 703.39 | 32 |
| complete | lite | 623.91 | 623.91 | 32 |
| complete | lite | 814.85 | 814.84 | 32 |
| complete | lite | 438.46 | 438.46 | 32 |
| complete | lite | 500.53 | 500.54 | 32 |
| complete | lite | 723.36 | 723.36 | 32 |
| complete | lite | 448.06 | 448.07 | 32 |
| complete | lite | 592.53 | 592.52 | 32 |
| complete | lite | 1106.87 | 1106.88 | 1024 |
| complete | lite | 1086.09 | 1085.99 | 1024 |
| complete | lite | 759.23 | 759.23 | 1024 |
| complete | lite | 1110.70 | 1110.66 | 1024 |
| complete | lite | 1206.49 | 1206.45 | 1024 |
| complete | lite | 1148.69 | 1148.69 | 1024 |
| complete | lite | 1135.75 | 1135.76 | 1024 |
| complete | lite | 1105.14 | 1105.12 | 1024 |
| complete | lite | 1236.00 | 1235.97 | 1024 |
| complete | lite | 1275.01 | 1274.94 | 1024 |
| complete | lite | 1039.72 | 1039.72 | 1024 |
| complete | lite | 1237.13 | 1237.13 | 1024 |
| complete | lite | 1218.68 | 1218.63 | 1024 |
| complete | lite | 1374.46 | 1374.21 | 1024 |
| complete | lite | 1310.89 | 1310.89 | 1024 |
| complete | lite | 1154.87 | 1154.87 | 1024 |
| complete | lite | 1085.11 | 1084.81 | 1024 |
| complete | lite | 834.39 | 834.38 | 1024 |
| complete | lite | 1010.97 | 1010.97 | 1024 |
| complete | lite | 1093.91 | 1093.91 | 1024 |
| complete | lite | 1382.04 | 1382.02 | 1024 |
| complete | lite | 1127.83 | 1127.83 | 1024 |
| complete | lite | 664.25 | 664.25 | 1024 |
| complete | lite | 1327.77 | 1327.77 | 1024 |
| complete | lite | 1113.74 | 1113.74 | 1024 |
| complete | lite | 1075.13 | 1075.13 | 1024 |
| complete | lite | 1121.92 | 1121.92 | 1024 |
| complete | lite | 1310.80 | 1310.79 | 1024 |
| complete | lite | 962.52 | 962.51 | 1024 |
| complete | lite | 1052.42 | 1052.42 | 1024 |
| complete | lite | 728.25 | 728.25 | 1024 |
| complete | lite | 835.02 | 835.02 | 1024 |
| complete | lite | 1325.92 | 1325.92 | 1024 |
| complete | lite | 1077.96 | 1077.96 | 1024 |
| complete | lite | 1241.28 | 1240.96 | 1024 |
| complete | lite | 1305.93 | 1305.91 | 1024 |
| complete | lite | 1049.91 | 1049.88 | 1024 |
| complete | lite | 1219.75 | 1219.71 | 1024 |
| complete | lite | 623.50 | 623.50 | 1024 |
| complete | lite | 989.61 | 989.61 | 1024 |
| complete | lite | 1023.78 | 1023.78 | 1024 |
| complete | lite | 1184.93 | 1184.81 | 1024 |
| complete | lite | 1061.89 | 1061.90 | 1024 |

| Library | Implementation | Wall Nanoseconds | CPU Nanoseconds | num agents |
|---|---|---|---|---|
| complete | lite | 1069.00 | 1068.97 | 1024 |
| complete | lite | 1267.68 | 1267.52 | 1024 |
| complete | lite | 1270.25 | 1270.25 | 1024 |
| complete | lite | 1210.18 | 1209.93 | 1024 |
| complete | lite | 1068.34 | 1062.88 | 1024 |
| complete | lite | 1166.18 | 1159.55 | 1024 |
| complete | lite | 1092.61 | 1087.22 | 1024 |
| complete | lite | 1177.96 | 1176.74 | 32768 |
| complete | lite | 1176.31 | 1176.20 | 32768 |
| complete | lite | 1026.71 | 1026.48 | 32768 |
| complete | lite | 1088.54 | 1086.15 | 32768 |
| complete | lite | 1155.79 | 1148.56 | 32768 |
| complete | lite | 1105.22 | 1103.75 | 32768 |
| complete | lite | 1090.18 | 1089.99 | 32768 |
| complete | lite | 1220.37 | 1220.18 | 32768 |
| complete | lite | 1020.25 | 1011.71 | 32768 |
| complete | lite | 1200.44 | 1192.92 | 32768 |
| complete | lite | 1144.20 | 1144.17 | 32768 |
| complete | lite | 1343.69 | 1343.68 | 32768 |
| complete | lite | 995.65 | 995.62 | 32768 |
| complete | lite | 1126.91 | 1126.89 | 32768 |
| complete | lite | 1011.32 | 1011.29 | 32768 |
| complete | lite | 1090.83 | 1090.77 | 32768 |
| complete | lite | 1093.61 | 1093.60 | 32768 |
| complete | lite | 1089.13 | 1089.12 | 32768 |
| complete | lite | 1246.96 | 1246.91 | 32768 |
| complete | lite | 1078.44 | 1078.44 | 32768 |
| complete | lite | 1261.88 | 1261.85 | 32768 |
| complete | lite | 1209.66 | 1209.65 | 32768 |
| complete | lite | 1236.44 | 1236.40 | 32768 |
| complete | lite | 953.64 | 953.65 | 32768 |
| complete | lite | 1086.47 | 1086.46 | 32768 |
| complete | lite | 1039.57 | 1039.50 | 32768 |
| complete | lite | 939.09 | 939.08 | 32768 |
| complete | lite | 970.19 | 970.14 | 32768 |
| complete | lite | 1154.79 | 1154.76 | 32768 |
| complete | lite | 1159.90 | 1159.85 | 32768 |
| complete | lite | 903.69 | 903.69 | 32768 |
| complete | lite | 958.45 | 958.07 | 32768 |
| complete | lite | 1047.56 | 1047.51 | 32768 |
| complete | lite | 959.05 | 959.04 | 32768 |
| complete | lite | 969.65 | 969.35 | 32768 |
| complete | lite | 1064.84 | 1064.82 | 32768 |
| complete | lite | 1057.66 | 1057.58 | 32768 |
| complete | lite | 1295.73 | 1295.73 | 32768 |
| complete | lite | 1119.69 | 1119.67 | 32768 |
| complete | lite | 1024.96 | 1024.93 | 32768 |
| complete | lite | 1029.08 | 1029.04 | 32768 |
| complete | lite | 898.97 | 898.92 | 32768 |
| complete | lite | 1310.38 | 1310.36 | 32768 |
| complete | lite | 1023.94 | 1023.92 | 32768 |
| complete | lite | 1211.95 | 1211.93 | 32768 |
| complete | lite | 1064.84 | 1064.81 | 32768 |

| Library | Implementation | Wall Nanoseconds | CPU Nanoseconds | num agents |
|---|---|---|---|---|
| complete | lite | 901.60 | 901.60 | 32768 |
| complete | lite | 958.30 | 958.29 | 32768 |
| complete | lite | 1207.84 | 1207.82 | 32768 |
| complete | lite | 1329.57 | 1329.54 | 32768 |
| nop | lite | 38.11 | 38.11 | 1 |
| nop | lite | 38.12 | 38.12 | 1 |
| nop | lite | 35.60 | 35.60 | 1 |
| nop | lite | 35.52 | 35.52 | 1 |
| nop | lite | 35.61 | 35.60 | 1 |
| nop | lite | 35.56 | 35.56 | 1 |
| nop | lite | 35.59 | 35.59 | 1 |
| nop | lite | 35.59 | 35.59 | 1 |
| nop | lite | 35.60 | 35.60 | 1 |
| nop | lite | 35.55 | 35.55 | 1 |
| nop | lite | 35.56 | 35.56 | 1 |
| nop | lite | 35.55 | 35.55 | 1 |
| nop | lite | 35.59 | 35.59 | 1 |
| nop | lite | 35.65 | 35.65 | 1 |
| nop | lite | 35.55 | 35.55 | 1 |
| nop | lite | 35.59 | 35.59 | 1 |
| nop | lite | 35.60 | 35.60 | 1 |
| nop | lite | 35.70 | 35.70 | 1 |
| nop | lite | 35.89 | 35.79 | 1 |
| nop | lite | 35.82 | 35.82 | 1 |
| nop | lite | 35.85 | 35.85 | 1 |
| nop | lite | 35.84 | 35.84 | 1 |
| nop | lite | 36.20 | 35.91 | 1 |
| nop | lite | 35.81 | 35.79 | 1 |
| nop | lite | 35.75 | 35.74 | 1 |
| nop | lite | 35.67 | 35.67 | 1 |
| nop | lite | 35.60 | 35.60 | 1 |
| nop | lite | 35.59 | 35.59 | 1 |
| nop | lite | 35.61 | 35.61 | 1 |
| nop | lite | 35.57 | 35.57 | 1 |
| nop | lite | 35.58 | 35.58 | 1 |
| nop | lite | 35.60 | 35.60 | 1 |
| nop | lite | 35.56 | 35.56 | 1 |
| nop | lite | 35.56 | 35.56 | 1 |
| nop | lite | 35.59 | 35.59 | 1 |
| nop | lite | 35.62 | 35.58 | 1 |
| nop | lite | 35.59 | 35.59 | 1 |
| nop | lite | 35.59 | 35.59 | 1 |
| nop | lite | 35.56 | 35.56 | 1 |
| nop | lite | 35.52 | 35.52 | 1 |
| nop | lite | 35.56 | 35.56 | 1 |
| nop | lite | 35.53 | 35.53 | 1 |
| nop | lite | 35.57 | 35.57 | 1 |
| nop | lite | 35.66 | 35.65 | 1 |
| nop | lite | 35.80 | 35.80 | 1 |
| nop | lite | 35.73 | 35.73 | 1 |
| nop | lite | 35.83 | 35.83 | 1 |
| nop | lite | 35.56 | 35.56 | 1 |
| nop | lite | 35.59 | 35.59 | 1 |

| Library | Implementation | Wall Nanoseconds | CPU Nanoseconds | num agents |
|---|---|---|---|---|
| nop | lite | 35.59 | 35.59 | 1 |
| nop | lite | 32.16 | 32.16 | 32 |
| nop | lite | 32.02 | 32.02 | 32 |
| nop | lite | 32.03 | 32.03 | 32 |
| nop | lite | 32.05 | 32.05 | 32 |
| nop | lite | 32.04 | 32.04 | 32 |
| nop | lite | 32.04 | 32.04 | 32 |
| nop | lite | 32.05 | 32.05 | 32 |
| nop | lite | 32.03 | 32.03 | 32 |
| nop | lite | 32.06 | 32.06 | 32 |
| nop | lite | 32.02 | 32.02 | 32 |
| nop | lite | 32.05 | 32.05 | 32 |
| nop | lite | 32.04 | 32.04 | 32 |
| nop | lite | 32.04 | 32.04 | 32 |
| nop | lite | 32.07 | 32.07 | 32 |
| nop | lite | 32.18 | 32.18 | 32 |
| nop | lite | 32.03 | 32.03 | 32 |
| nop | lite | 32.07 | 32.07 | 32 |
| nop | lite | 32.08 | 32.08 | 32 |
| nop | lite | 32.08 | 32.08 | 32 |
| nop | lite | 32.10 | 32.09 | 32 |
| nop | lite | 32.06 | 32.06 | 32 |
| nop | lite | 32.01 | 32.01 | 32 |
| nop | lite | 32.09 | 32.09 | 32 |
| nop | lite | 32.01 | 32.01 | 32 |
| nop | lite | 32.05 | 32.05 | 32 |
| nop | lite | 32.05 | 32.05 | 32 |
| nop | lite | 32.03 | 32.03 | 32 |
| nop | lite | 32.05 | 32.05 | 32 |
| nop | lite | 32.01 | 32.01 | 32 |
| nop | lite | 32.05 | 32.05 | 32 |